\documentclass[letterpaper, 10 pt, conference]{ieeeconf}  

\IEEEoverridecommandlockouts                              

\usepackage{amsmath}
\usepackage[super]{nth}
\usepackage{amsmath,bm}
\usepackage{booktabs}
\usepackage{graphicx}
\usepackage{caption}
\usepackage{subcaption}
\usepackage{gensymb}
\usepackage{mathtools}

\DeclarePairedDelimiter\floor{\lfloor}{\rfloor}
\usepackage{hyperref}
\hypersetup{
    colorlinks=true,
    linkcolor=red,
    filecolor=magenta,      
    urlcolor=magenta,
}
\usepackage{cite}

\newcommand{\vect}[1]{\mathbf{#1}}
\newcommand{\Lagr}{\mathcal{L}}
\newcommand{\mypara}[1]{\par\vspace*{1.5mm}\noindent\textbf{{#1}}}

\title{\LARGE \bf
Visual Perspective Taking for Opponent Behavior Modeling
}

\author{Boyuan Chen* \quad Yuhang Hu \quad Robert Kwiatkowski \quad Shuran Song \quad Hod Lipson\\
Columbia University
\thanks{We thank Philippe Wyder and Carl Vondrick for helpful discussions. We thank all the reviewers for their great comments and efforts. This research is supported by NSF NRI 1925157 and DARPA MTO grant HR0011-18-2-0020. *bchen@cs.columbia.edu}
}

\begin{document}

\maketitle
\thispagestyle{empty}
\pagestyle{empty}

\begin{abstract}

In order to engage in complex social interaction, humans learn at a young age to infer what others see and cannot see from a different point-of-view, and learn to predict others' plans and behaviors. These abilities have been mostly lacking in robots, sometimes making them appear awkward and socially inept. Here we propose an end-to-end long-term visual prediction framework for robots to begin to acquire both these critical cognitive skills, known as Visual Perspective Taking (VPT) and Theory of Behavior (TOB). We demonstrate our approach in the context of visual hide-and-seek – a game that represents a cognitive milestone in human development. Unlike traditional visual predictive model that generates new frames from immediate past frames, our agent can directly predict to multiple future timestamps (25 s), extrapolating by 175\% beyond the training horizon. We suggest that visual behavior modeling and perspective taking skills will play a critical role in the ability of physical robots to fully integrate into real-world multi-agent activities. Our website is at \url{http://www.cs.columbia.edu/~bchen/vpttob/}.

\end{abstract}

\section{INTRODUCTION}

Imagine that you see your colleague trying to open the door, only to find that the door is blocked by the doorstop hidden on the other side. Without exchanging words, you rush to remove the doorstop, allowing the door to swing open. You took this action because you understood what your colleague was trying to do (open the door), and you understood that she could not see the doorstop (from her viewpoint). We refer to these intuitive abilities as Behavior Modeling and Visual Perspective Taking respectively. Visual Perspective Taking \cite{piaget2013child, flavell1977development, flavell1980young, newcombe1992children, moll2011does, pearson2013review} is essential for many useful multi-agent skills such as understanding social dynamics and relationships, engaging in social interactions, and understanding intentions of others. A simple action such as getting out of the way of a busy person, or assisting a person about to sit down, involve both of these abilities.

\begin{figure}[t]
\begin{center}
    \includegraphics[width=.475\textwidth]{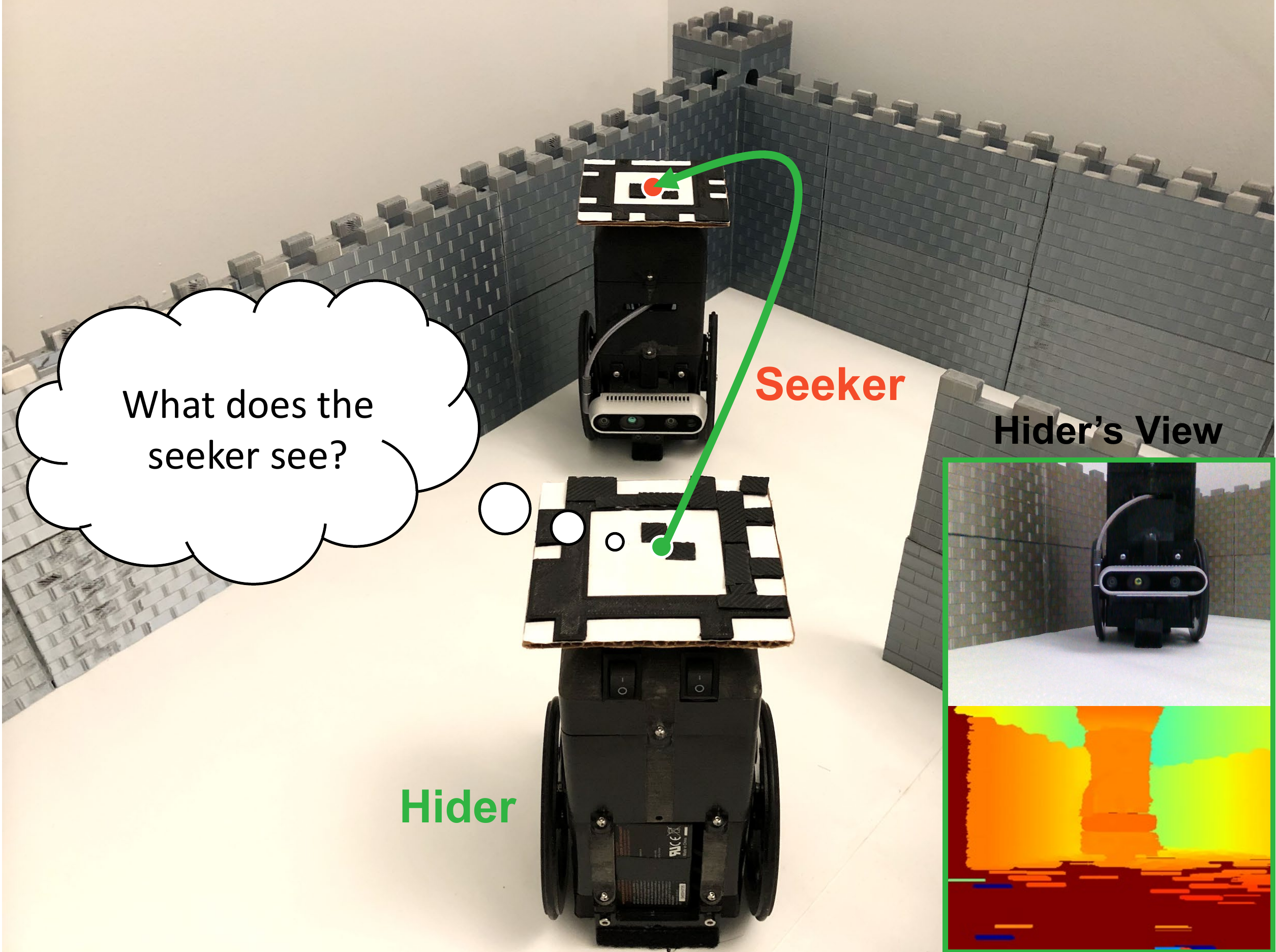}
\end{center}
\vspace{-3mm}
\caption{Visual Perspective Taking (VPT) refers to the ability to estimate other agents' viewpoint from its own observation. With our VPT framework, the hider robot learns to predict the seeker robot future observation and plan its actions based on this prediction to avoid being captured.}
\label{fig:teaser}
\vspace{-13pt}
\end{figure}

Like humans, robots that can perform VPT and use that ability to model others \cite{chen2021visual} have many advantages across a variety of applications such as social robotics, assistive and service robotics. However, even though current robots can perform variety of tasks by themselves or move as a group, these robots often do not demonstrate the capability of performing perspective taking or behavior modeling. Several challenges still remain.

First, visual observations are high-dimensional with large amount of information. Relying on manual feature extractions and hand-designed learning procedure \cite{trafton2006children} cannot scale in complex setup. Second, the behaviors among different robots are cross-dependent. Consequently, it is necessary to account for the mutual interaction \cite{albrecht2018autonomous} between related robots. Third, directly learning on physical robots requires strong data efficiency. Though recent methods \cite{chen2020visual, baker2020emergent, weihs2019artificial, labash2020perspective} demonstrate that similar cognitive abilities could emerge from reward-driven learning, they often require millions of learning steps. Lastly, learning to predict long-term future is very challenging. Most existing visual prediction frameworks \cite{leibfried2016deep, finn2017deep, racaniere2017imagination, chiappa2017recurrent, hafner2019learning} follow an iterative paradigm to rollout the prediction which are both time and memory consuming under long-horizon tasks.

In this paper, we propose a self-supervised vision-based learning framework to take a small step towards realizing VPT and TOB in navigation robots involving future opponent modeling. We call our framework VPT-TOB. Our key idea is to explicitly predict the future perspective of the other robot as an image by conditioning on a top-down visual embedding of first-person RGB-D camera images and time-abstracted action embeddings. We further present a value prediction model that can leverage the future perspective prediction to evaluate action proposals. During test time, we ask the robot to propose future goals on the top-down map and generates potential action plans with its low-level controller. The robot can then use our VPT-TOB model to imagine future perspectives of the other robot and evaluates its goal proposal with the value prediction model. The robot can choose the safest position to navigate to.

Our experiments on a physical hide-and-seek task suggest that our method outperforms baseline algorithms which do not explicitly consider other agents' perspectives and behaviors, showing the importance of explicit visual perspective taking and behavior modeling. Our hider robot exhibits diverse behaviors during hiding and an interpretable decision-making process by providing how the robot is envisioning the future state and actions of another robot.

The primary contribution of this paper is to demonstrate that a long-term vision prediction framework can be used to model the perspectives and behaviors of other robots. We present a value prediction module that can take the perspective prediction to evaluate action proposals during test time. We perform several experiments and ablation studies in both real and simulation environments to evaluate the effectiveness of our design decisions.

\section{RELATED WORKS}

We chose to demonstrate and evaluate the capacity of robots to perform perspective taking and behavior modeling in the game of hide-and-seek. Hide-and-seek has been a representative example \cite{pearson2013review, street2018perspective} in Cognitive Science and Psychology to study children’s development of VPT skills. Hide-and-seek has also been suggested \cite{reinhold2019behavioral} as a promising framework to study VPT and Theory of Mind related abilities in animals. In order to demonstrate robots with VPT and ToB abilities, we build a simple visual embodied environment to play a similar hide-and-seek game between two robots.

In Embodiment AI and Robotics, earlier works \cite{trafton2006children, trafton2005enabling} present a cognitive architecture for a robot to play hide and seek with humans by learning from a hand-designed rules and feature extraction. Recently, Chen et al. \cite{chen2020visual} proposes a first-person visual hide-and-seek task and analyzes the emerged behaviors under various environment conditions. Baker et al. \cite{baker2020emergent} trains the agents to play the game on low-dimensional observations and found that more complex tool usage behaviors can emerge. Weihs et al. \cite{weihs2019artificial} observes similar behaviors by training the agents in a variant of hiding task with objects. There are also significant works \cite{bratman2010new, foerster2016learning, lazaridou2016multi, grouchy2016evolutionary, sukhbaatar2016learning, mordatch2017emergence, cao2018emergent, jain2019two} on multi-agent communications with carefully designed protocols. However, these works only operate in simulation environments with perfect sensory inputs. These approaches also require at least millions of learning steps, which make it impractical for real-world robotic applications. Our work, on the other hand, trains the physical robot from scratch while also showing diverse learned behaviors.  

Our work is also closely related to model-based robot learning from visual inputs. There have been significant recent research on learning to plan with Model Predictive Control (MPC) by first learning the dynamics from visual inputs ranging from object manipulation \cite{finn2017deep, byravan2018se3, hoque2020visuospatial}, locomotion \cite{hafner2019dream, hafner2019learning}, and navigation \cite{ha2018world, hirose2019deep}. A typical setup is to train a model to predict future states from the current and past observations, and then the same model or a separate model is used to estimate the state-action value for goal-driven planning or reward-driven policy learning. Our work differs from these studies mainly on two aspects. First, instead of predicting the future in an auto-regressive manner, our method can directly envision the future within any step in the horizon with one-step rollout. Second, our work focuses on opponent's perspective and behavior modeling while other works research single-agent skill learning.

\section{METHOD OVERVIEW}

In this paper, we frame the Visual Perspective Taking and Behavior Modeling as an action-conditioned vision predictive model. Our method encourages the learning robot to explicitly contemplate the future visual perspective and behavior of the other robot by outputting where the other agent will be, and what the other agent will see as an image. Our hider robot can be trained directly on the physical setup. Once trained, the robot can answer “what if” questions visually: what will the other agent do, and what will its view become, if I choose to perform this action for some time? 

\begin{figure*}[!t]
    \vspace*{5pt}
    \centering
    \includegraphics[width=0.975\linewidth]{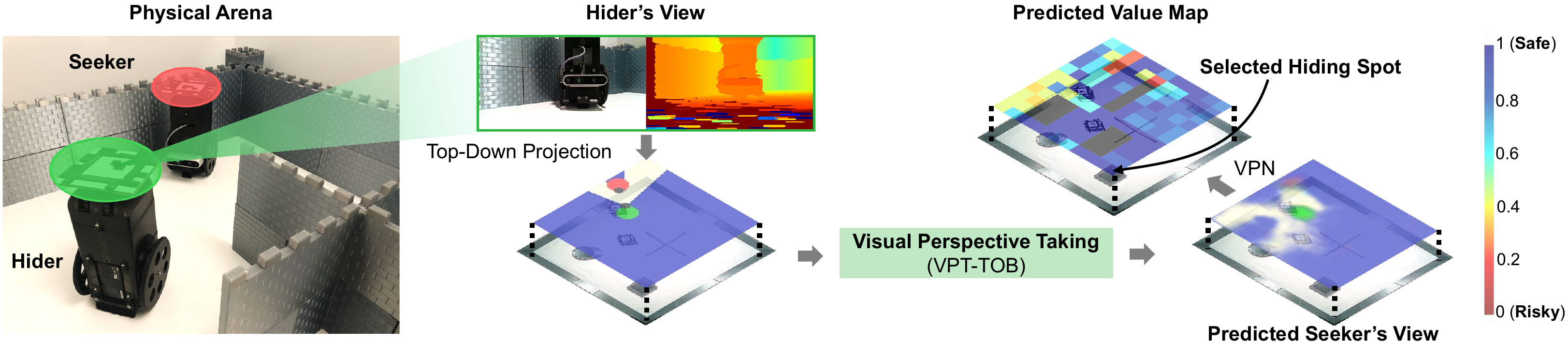}
    \caption{\textbf{Method Overview:} our robot learns to infer the future view of the other robot with our VPT-TOB model based on its initial visual observation and future action embeddings (blue is invisible area). With these anticipations and a value prediction model, the robot can produce a value map indicating the safety level (color bar) and selects the best hiding spot.}
    \label{fig:overview}
    \vspace{-13pt}
\end{figure*}

\mypara{Task Setup} To investigate the effectiveness of this framework in opponent modeling, we consider a hide-and-seek setting \cite{chen2020visual} where a hider robot needs to avoid being captured by another seeker robot while navigating around the environment with various obstacles. In order to predict the future perspective of the seeker, the hider robot needs to infer the geometric view from the seeker, reason about the environment spatial layout, and form a model of the seeker's policy. This task reflects several challenging aspects in real-world setup such as occlusions in robot perception and cross-dependent actions among multiple robots.

We 3D printed two-wheeled robots with size of $110$mm by $130$mm on a 1.2m by 1.2m plane with three obstacles. Similar to Wu et al. \cite{wu2020spatial}, for simple prototype, we place fiducial markers \cite{olson2011apriltag} on the robots and track them with a top-down camera. However, our setup aims at representing what could be possible to achieve with only the onboard RGB-D camera for SLAM. Therefore, the robots only perceive what could be captured with the first-person RGB-D camera with 86\degree FoV with noisy depth and color information and occlusions.

Both robots start by facing at each other at a random location in the environment and moves in the same speed (10 mm/step). The robots can move forward or backward and rotate with 10\degree as given primitives. We only focus on stationary case where the seeker robot follows a heuristic expert policy and leave the non-stationary scenario for future work as the first step towards grounding VPT to physical robots. If the hider is visible through its first-person camera, the seeker will navigate towards it with A* \cite{hart1968formal}; if the seeker loses the hider in its camera view, it will navigate to the last known position of the hider and then continuously explore the room until the hider shows again.

\subsection{State Representation}

We represent the robot state observation as a RGB image. The image representation depicts a top-down visibility map. This is similar to the representations in other mobile navigation and self-driving systems \cite{gao2017intention, chen2018deep, shah2018follownet, bruls2019right, wu2020spatial}. We project the depth information from first-person view to a 2D bird's eye view to acquire the top-down visibility map. Areas behind the object boundary are treated as free space. We colorize the observation map to reflect this. The hider robot is always aware of its own position, while the seeker position is only available when it is visible through a color filter from the first-person view. The robots are plotted as circles with different colors. We denote the final visual state representation of the hider and the seeker at timestamp t as $I_{H, t}$ and $I_{S, t}$.

\subsection{Action Representation}

\begin{figure}[t]
\begin{center}
    \includegraphics[width=.475\textwidth]{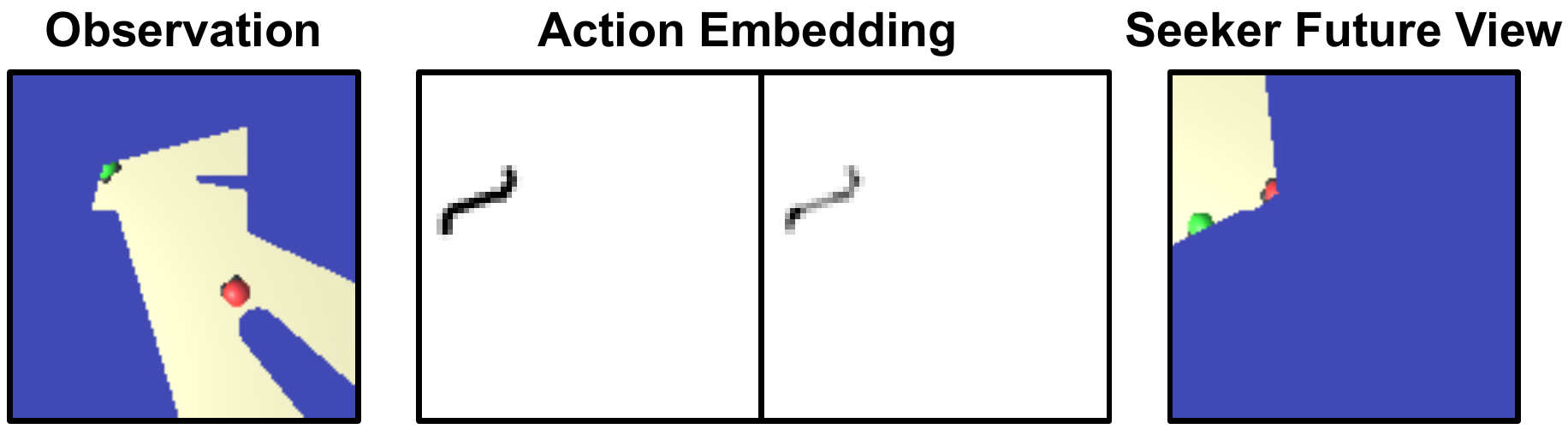}
\end{center}
\vspace{-3mm}
\caption{An example of the training data pair for VPT-TOB.}
\label{fig:datasample}
\vspace{-18pt}
\end{figure}

We design the action representation with two key considerations in mind. First, we want to align the action representation with the state representation. Second, we need to preserve information about past action sequences. Thus, we propose to encode the robot's trajectory sequence as two 2-channel images with the same dimensions as our top-down visibility map. Say we are interested in the action sequence from $t{=}0$ up to a future time $t{=}t_i$. The first image is a visitation map $F_{t{=}0:t_i}$ denoting how many times each point on the map will be visited by the robot. This is achieved through the second alpha-channel where darker color represents more frequent visitation. The second image is a time-encoded trajectory map $T_{t{=}0:t_i}$ denoting the traversal order. Similar to $F$, the darker the color in $T$, the later the position will be traversed. Both $F$ and $T$ are necessary for a complete action representation to avoid the ambiguity brought by the trajectory intersection. An example of a training data pair is shown in Fig. \ref{fig:datasample}.

\subsection{Data Collection}

We experiment with two types of exploration policies for the hider to collect the training data. The first is a random policy created by sampling from the given action primitives. However, this policy explores poorly thus resulting in the seeker easily catching the hider. The hider mostly stays alive for only 20 to 30 steps which biases the training data to be short-sighted. The second policy is human play data \cite{lynch2020learning}. We ask a human subject to play as the hider in front of a display from the first-person view for $100$ trajectories. Since the human subject has never played this version of the game before, the trajectories are diverse ranging from successful trials to failures, making it a good strategy to stay with.

During robot data collection, the hider robot randomly samples a human trajectory and uses the selected trajectory as is but from a different initial condition. It is also possible to train an imitation policy on the human data, but we find the current strategy sufficient to give enough balanced data. A horizon will terminate when one of two conditions is triggered: 1) the seeker robot catches the hider robot, or 2) the total number of steps reaches a maximum value ($100$ in our setup).

\subsection{VPT-TOB Network}

The objective of the VPT-TOB network $f_p$ is to anticipate the visual observation $\vect{I_{S, t{=}t_i}}$ of the seeker at future time step $t_i$ given its initial observation $\vect{I_{H, t{=}0}}$ and action embedding ($\vect{F_{t{=}0:t_i}}$, $\vect{T_{t{=}0:t_i}}$). Intuitively, the hider robot needs to first anticipate where it will end up if it takes certain action sequences. Secondly, the robot needs to model the behavior of the other robot so that it learns how the seeker will react. Finally, the hider needs to project the view using its prediction of the seeker agent, which requires an understanding of scene geometry and view projection. During training, the hider has access to the seeker's view as a supervised signal. During testing, the hider has to predict the seeker's views.

\mypara{Network Architecture} Our VPT-TOB model is a fully convolutional encoder-decoder network \cite{chen2021visual}. The encoder is a 8-layer fully convolutional network which takes in the concatenated visual observation and action embedding as input with size 128 $\times$ 128. The decoder network comprises four transposed convolutional layers where each of them is followed by another 2D convolution and a transposed convolution for high-resolution output \cite{dosovitskiy2015flownet} with size $128 \times 128$.

\mypara{Training} VPT-TOB model can be supervised with a simple pixel-wise mean-squared error loss. We use 1,000 real-world trajectories for training and 200 trajectories for validation and testing. We use Adam \cite{kingma2014adam} optimizer with a learning rate of $0.001$ and batch size $256$. We decrease the learning rate by $10\%$ at the $25\%$ and $65\%$ of the training progress. Formally, the loss function can be written as:
\begin{equation*}
    \Lagr_{\text{VPT-TOB}} = \text{MSE}(f_p(\vect{I_{H, t{=}0}}, \vect{F_{t{=}0:t_i}}, \vect{T_{t{=}0:t_i}}), \vect{I_{S, t{=}t_i}})
\end{equation*}

\subsection{Inference and Planning}

At test time, the hider robot needs to propose a set of action plans and envision the possible outcomes for them with VPT-TOB to pick the best plan. One way to represent the plan is to specify a goal location and then perform visual MPC \cite{finn2017deep, ebert2018visual} with the learned opponent model to find the best path. However, in our task setup in which no human specified goal is available, a more reasonable way is to propose a spatial location in the room and plan with a low-level controller policy (\textit{e.g.}, A*). This parameterization emphasizes a safer hiding location instead of a safer path to follow given a required goal.

\mypara{Value Estimation Map} We therefore aim to generate a value map to evaluate the risk level of the possible goal locations. A value map is a matrix where each element denotes the safety of choosing the corresponding position as the goal. The earlier the hider robot is caught towards that goal, the riskier that goal is. Formally, the value for each element on the value map is defined as the following accumulation form:
\begin{align*}
    V_{i, t{=}0:T} = \sum_{k} v_{i, t=(kN-1)},\ 
    v = 
    \begin{cases}
        1, & \text{if hider is caught}\\
        0, & \text{otherwise}
    \end{cases}
\end{align*}
where $k$ is an integer from $0$ to $\floor*{\frac{T}{N}}$ and $T$ is the maximum time budget for this horizon and $N$ is the step size. Under this formulation, we discretize the environment as a grid map (121 endpoints for our setup). From the current hider position, the planned path from A* can be embedded every N (N{=}10 for us) steps.

\mypara{Value Prediction Network (VPN)} Our value prediction network $f_v$ takes in the visual prediction from VPT-TOB to output the binary $v$ value for each step. We can then use the above formula to compute the accumulated value along all the steps. We just need a few forward passes on VPT-TOB and VPN to obtain the entire value map within $1.4$ seconds on a single GPU.

\begin{figure*}[t]
    \vspace*{5pt}
    \centering
    \includegraphics[width=0.98\linewidth]{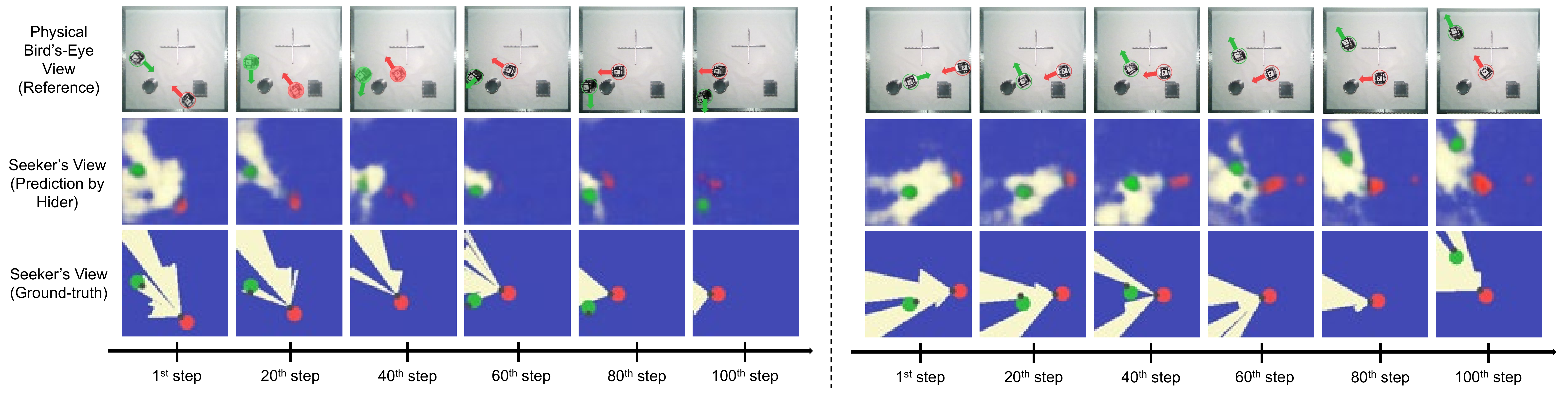}
    \caption{\textbf{VPT-TOB predictions from physical robots:} we here show two example sequences. The predictions are shown as a function of time. Our hider robot accurately predicts the future perspectives of the seeker robot only by its own initial observation and a future potential action plan.}
    \label{fig:real-vpttob}
    \vspace{-15pt}
\end{figure*}

Intuitively, VPN estimates whether the hider will be caught given the predicted future view of the seeker. Our VPN network consists of 3-convolutional layers followed by 2 fully-connected layers. We apply max-pooling after each convolution and a dropout layer right after the first fully-connected layer. The network is trained to minimize a cross-entropy loss. We augment the data by randomly rotating the images among $(90\degree, 180\degree, 270\degree)$. In total, we have $2,000$ images for training, and $400$ images each for validation and evaluation. We train our model for $100$ epochs with a batch size of $256$ and a learning rate of $0.0005$. We decrease the learning rate by $10\%$ at $20\%$ and $50\%$ of the training progress. We use Adam as the optimizer. Formally, the loss function can be written as:
\begin{equation*}
    \Lagr_{\text{VPN}} = -(v\log(f_v(\vect{\hat{I}_{S, t{=}t_i}}))) + (1-v)\log(1-f_v(\vect{\hat{I}_{S, t{=}t_i}}))
\end{equation*}

\section{EXPERIMENTS}

We first perform a series of experiments in our real-world setup. To evaluate different components in our algorithm, we further replicate our physical setup in a simulation and execute several ablation and baseline comparisons.

\subsection{Evaluation Metrics} \label{sec: metric}

\mypara{VPT-TOB and VPN} A straightforward evaluation of the VPT-TOB model is to provide quantitative visualizations of the imagined frames. Since our goal with VPT-TOB is to capture the perspective information and opponent behavior for long-term planning, a perceptual error will not be ideal for our evaluation. Meanwhile, our VPN model needs to take in the predicted frame from VPT-TOB to infer whether the hider robot will be safe. Therefore, we can use the performance of VPN on the output of VPT-TOB as a quantitative metric for our VPT-TOB model.

\mypara{Value Estimation Map} To evaluate the overall performance of the entire planning pipeline, we quantify the accuracy with the entire value map. We measure this in simulation which can provide us with ground-truth values conveniently by moving the hider robot to all possible goal locations and collecting the accumulated values. During test time, what matters for decision making is the relative value between different goal locations instead of the absolute value. To this end, our metric for the value map is a relative ranking accuracy. Say we choose $g_i$ and $g_j$ as a pair to evaluate. The relative ranking of this pair is correct if it matches the relative ranking from the ground-truth value map. We evaluated all pairs at least 3 units away. This metric will result in the same ranking among all comparisons as the last metric, but provide an overall evaluation.

\subsection{Real-World Results}

\begin{figure}[t]
\begin{center}
    \includegraphics[width=.475\textwidth]{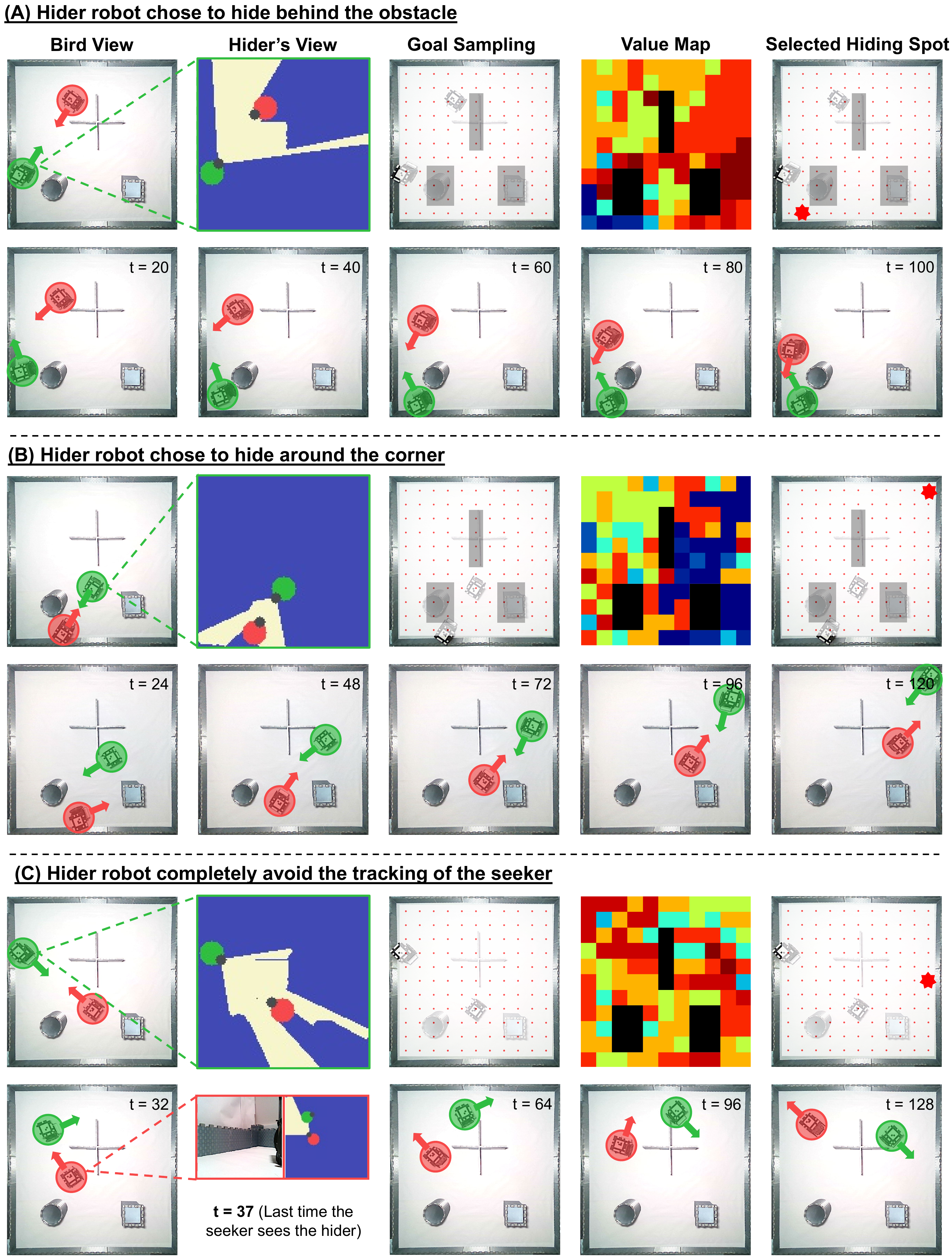}
\end{center}
\vspace{-3mm}
\caption{\textbf{Planning and hiding} Our hider robot produces a value map with our VPT-TOB and VPN networks. (Blue is safe and red is dangerous.) The hider then navigates to the farthest location with the highest value (star mark). Interestingly, in the last example, the hider completely got rid of the seeker from the \nth{37} time step.}
\label{fig:real-run}
\vspace{-13pt}
\end{figure}

The prediction results from our VPT-TOB model is a sequence of images based on a single initial observation from the hider's view and the future action embedding. In Fig. \ref{fig:real-vpttob}, we show the predictions from the random hider robot every $20$ steps up to $100$ steps corresponding to about $25$ seconds. Each predicted image presents where the hider thinks the seeker robot will be, and what the seeker's view will look like. Our model gives accurate predictions about the perspective and behavior of the opponent robot.

For test time planning, we randomly place the hider and seeker robot in the environment and ask the hider robot to plan and execute its goal selection from VPT-TOB and VPN. Fig. \ref{fig:real-run} shows the video frames where the hider robot navigates to its selected goal and successfully avoids the seeker only by relying on its initial value map estimation for the future 100 time steps. There is no extra decision making cost during execution. Interesting hiding behaviors emerged automatically from our setup as shown in Figure \ref{fig:real-run}.

We further test the performance of our VPN model under the real-world setup by inputting the predictions from the VPT-TOB model. Our VPN model achieves an accuracy of $88.39\%$. As we will discuss in the next section, this is very similar to what we can achieve in the simulation, demonstrating the effectiveness of our method in the physical robot platform.

\subsection{Simulation Analysis}

Our simulation is built with Unity \cite{juliani2018unity} to evaluate different components in our approach. The simulation can run in parallel and hence provides an efficient and scalable testing environment. To provide extensive analysis, we use in total 4,500 trajectories as training and 500 trajectories each for validation and testing. We can also get the top-down observation directly by applying a mesh filter in Unity with similar result from our real robot sensors. We run the hider robot for $200$ steps.

\subsection{Baseline Methods}

\mypara{Self-perspective} predicts the hider’s own future states conditioned on its initial observation and action plans other than explicitly modeling the other robot. This follows recent works on learning visual world models \cite{finn2017deep, racaniere2017imagination, byravan2018se3, hafner2019learning, hafner2019dream} with adapted architectures for fair comparison.

\mypara{Coordinate-value} directly predict whether the hider robot will be caught in the future from initial observation and action plans. The input consists of an initial state observation from the hider, a vector action embedding with spatial coordinate information, and a future coordinate value where the hider will end up using the given actions.

\mypara{Vector-action} is the same with VPT-TOB but replaces our action representation with a vector embedding as in \cite{buesing2018learning}. This embedding is concatenated with the intermediate features to produce the final predicted image.

The self-perspective baseline aims to test whether modeling the opponent agent is necessary. Both the self-perspective and coordinate-value baselines serve to examine if explicit opponent modeling is more effective than relying on the network to learn the opponent model implicitly. Vector-action is used to evaluate if our action representation can improve the quality of VPT-TOB model.

\begin{table}[h]
\vspace{-3pt}
\caption{VPT-TOB and VPN Performance}
\label{tab:vpn-baselines}
\centering
\resizebox{0.9\columnwidth}{!}{
\begin{tabular}{@{}lc@{}}
\toprule
\textbf{Method}         & \textbf{Success Rate on Held-out Examples} \\ \midrule
Self-perspective + VPN        & 76.61\%               \\
Coordinate-value        & 76.40\%               \\ \midrule
Vector-action + VPN           & 82.98\%               \\
\textbf{VPT-TOB (ours)} & \textbf{88.45\%}      \\ \bottomrule
\end{tabular}
}
\vspace{-3pt}
\end{table}

\subsection{Baseline Comparisons}

Through our simulation comparisons, we show the accuracy of VPN across all the applicable methods in Tab. \ref{tab:vpn-baselines} in which the input of VPN comes from our VPT-TOB or other baseline models. We run all the methods on 1,000 unseen examples with uniform starting locations and horizon length. The result suggest the advantage of our method. Both VPT-TOB and vector-action outperforms self-perspective and coordinate-value, validating the necessity to explicitly model the opponent perspective and behavior. The gap between vector-action and VPT-TOB indicates that our action representation offers additional gains for learning an accurate visual opponent model.

\begin{figure}[t]
\vspace*{2pt}
\begin{center}
    \includegraphics[width=.475\textwidth]{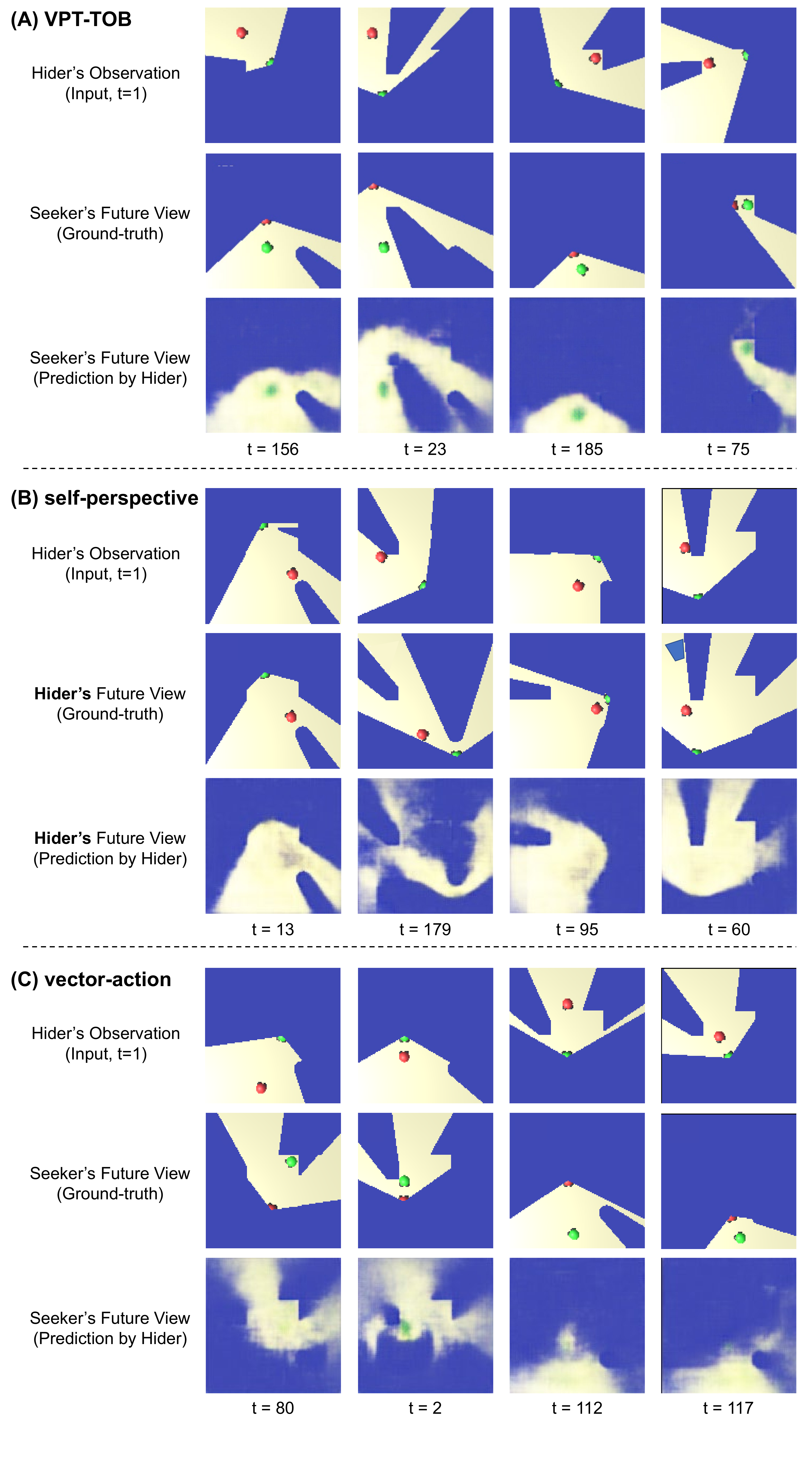}
\end{center}
\vspace{-3mm}
\caption{Predictions from VPT-TOB and other baselines. Our method produces more accurate modeling of the other robot's future perspective, suggesting the advantage of explicit opponent modeling and our visual action embedding.}
\label{fig:baseline-comparison}
\vspace{-13pt}
\end{figure}

Visualizations of the predicted frames substantiate our conclusion (Fig. \ref{fig:baseline-comparison}). For better comparison, we also display bird's-eye view in the first row and ground-truth seeker's view in the last row. However, these images are only for visualization, not inference.

\begin{figure}[t]
\vspace*{2pt}
\begin{center}
    \includegraphics[width=.475\textwidth]{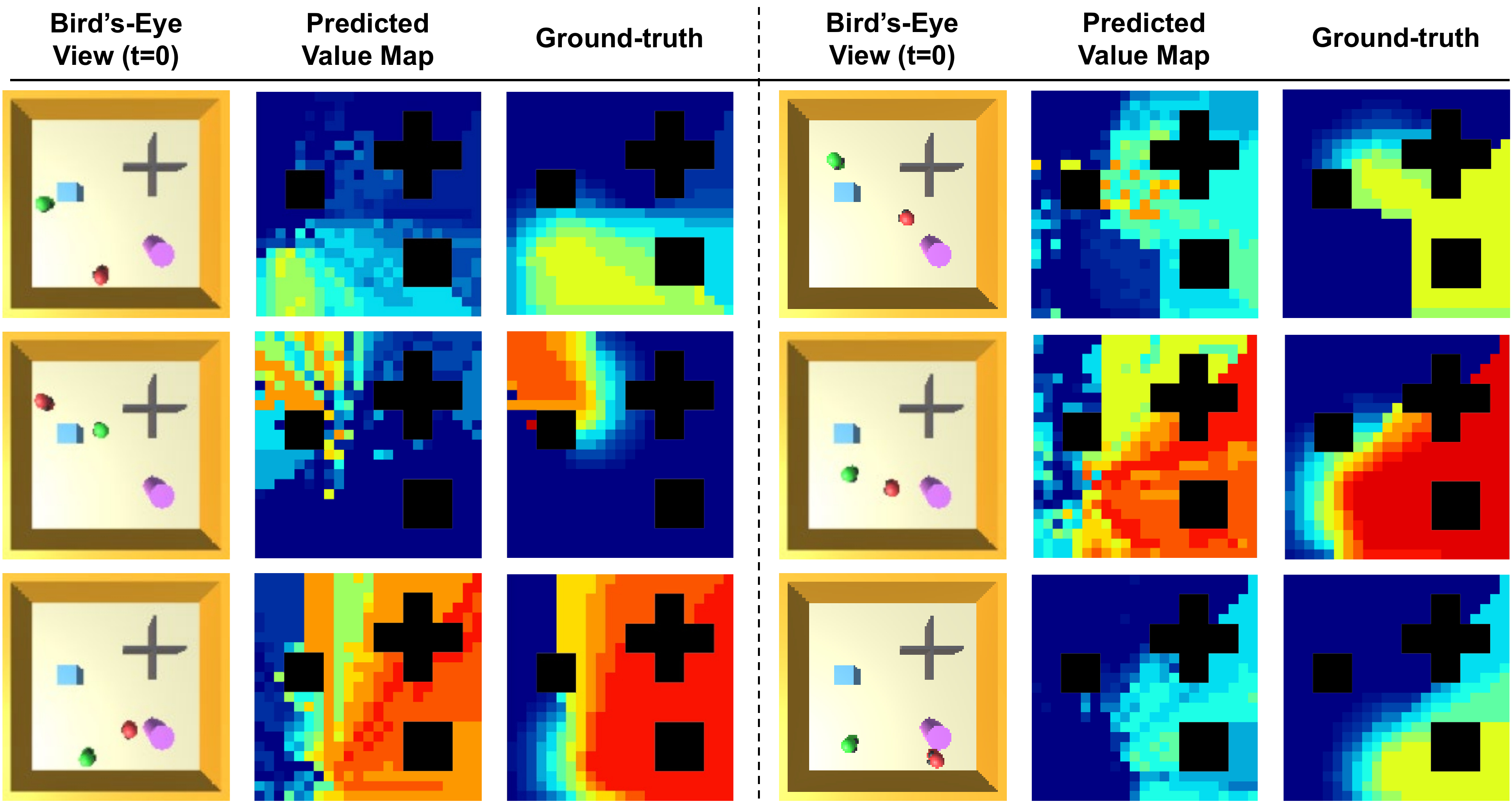}
\end{center}
\vspace{-3mm}
\caption{Predicted value maps versus ground-truth value maps. Our method produces accurate value estimations and reflect the overall value patterns.}
\label{fig:value-map-compare}
\vspace{-13pt}
\end{figure}

Fig. \ref{fig:value-map-compare} shows the generated value maps from our approach comparing with the ground-truth value maps. Overall, our framework demonstrates an accurate estimation about the safety level across the room.

\subsection{Generalization Across Training Time Horizon} With the relative ranking accuracy, we can quantify the performance of our approach from a higher-level perspective. Since our model should also learn temporal abstractions from our action representation, we extend the test horizon up to $175\%$ of the maximum number of steps given in training and examine how the relative ranking accuracy changes. We achieved this test by simply projecting longer horizon action trajectories on our image-based action embedding.

Tab. \ref{tab:prediction-horizon} shows the performance. Our model is able to predict far beyond its training horizon with very small losses of the accuracy. This suggests that our method also learns to model the temporal dynamics while modeling the other agent.

\begin{table}[h]
\vspace{-3pt}
\caption{Generalization on Prediction Horizon}
\label{tab:prediction-horizon}
\centering
\resizebox{0.9\columnwidth}{!}{
\begin{tabular}{@{}lc@{}}
\toprule
\textbf{Prediction Steps}      & \textbf{Relative Ranking Accuracy} \\ \midrule
200 (max seen during training) & 75.29 $\pm$ 12.23\%                  \\
250                            & 73.75 $\pm$ 10.35\%                  \\
300                            & 73.03 $\pm$ 10.40\%                  \\
350                            & 72.77 $\pm$ 10.82\%                  \\ \bottomrule
\end{tabular}
}
\vspace{-6pt}
\end{table}

\section{CONCLUSIONS AND FUTURE WORK}

We propose a visual predictive framework for realizing Visual Perspective Taking and Behavior Modeling in real-world robotic system. Our approach shows how to explicitly model the opponent's view and use the predictions for long-term planning. Our work takes a  step towards a general implementation of VPT and TOB ability on physical robots.

Our work studies VPT and TOB in a two-robot adversarial task. An interesting future direction is to explore the generalization to more robots. Additionally, our current method assumes a fixed policy for the seeker robot. It would be a useful next step to incorporate scenarios where the seeker policy might change by considering stochastic optimization with few-shot learning algorithms in an iterative learning manner. Finally, it would be an interesting topic to study swarm robot \cite{chung2018survey, gs2018survey} applications to deepen our understanding about more complex social dynamics.

\bibliographystyle{IEEEtran}
\bibliography{IEEEabrv, mybibfile}

\begin{thebibliography}{10}
\providecommand{\url}[1]{#1}
\csname url@rmstyle\endcsname
\providecommand{\newblock}{\relax}
\providecommand{\bibinfo}[2]{#2}
\providecommand\BIBentrySTDinterwordspacing{\spaceskip=0pt\relax}
\providecommand\BIBentryALTinterwordstretchfactor{4}
\providecommand\BIBentryALTinterwordspacing{\spaceskip=\fontdimen2\font plus
\BIBentryALTinterwordstretchfactor\fontdimen3\font minus
  \fontdimen4\font\relax}
\providecommand\BIBforeignlanguage[2]{{%
\expandafter\ifx\csname l@#1\endcsname\relax
\typeout{** WARNING: IEEEtran.bst: No hyphenation pattern has been}%
\typeout{** loaded for the language `#1'. Using the pattern for}%
\typeout{** the default language instead.}%
\else
\language=\csname l@#1\endcsname
\fi
#2}}

\bibitem{piaget2013child}
J.~Piaget, \emph{Child's Conception of Space: Selected Works vol 4}.\hskip 1em
  plus 0.5em minus 0.4em\relax Routledge, 2013, vol.~4.

\bibitem{flavell1977development}
J.~H. Flavell, ``The development of knowledge about visual perception.'' in
  \emph{Nebraska symposium on motivation}.\hskip 1em plus 0.5em minus
  0.4em\relax University of Nebraska Press, 1977.

\bibitem{flavell1980young}
J.~H. Flavell, E.~F. Flavell, F.~L. Green, and S.~A. Wilcox, ``Young children's
  knowledge about visual perception: Effect of observer's distance from target
  on perceptual clarity of target.'' \emph{Developmental Psychology}, vol.~16,
  no.~1, p.~10, 1980.

\bibitem{newcombe1992children}
N.~Newcombe and J.~Huttenlocher, ``Children's early ability to solve
  perspective-taking problems.'' \emph{Developmental psychology}, vol.~28,
  no.~4, p. 635, 1992.

\bibitem{moll2011does}
H.~Moll and A.~N. Meltzoff, ``How does it look? level 2 perspective-taking at
  36 months of age,'' \emph{Child development}, vol.~82, no.~2, pp. 661--673,
  2011.

\bibitem{pearson2013review}
A.~Pearson, D.~Ropar, A.~F.~d. Hamilton, \emph{et~al.}, ``A review of visual
  perspective taking in autism spectrum disorder,'' \emph{Frontiers in human
  neuroscience}, vol.~7, p. 652, 2013.

\bibitem{chen2021visual}
B.~Chen, C.~Vondrick, and H.~Lipson, ``Visual behavior modelling for robotic
  theory of mind,'' \emph{Scientific Reports}, vol.~11, no.~1, pp. 1--14, 2021.

\bibitem{trafton2006children}
J.~G. Trafton, A.~C. Schultz, D.~Perznowski, M.~D. Bugajska, W.~Adams, N.~L.
  Cassimatis, and D.~P. Brock, ``Children and robots learning to play hide and
  seek,'' in \emph{Proceedings of the 1st ACM SIGCHI/SIGART conference on
  Human-robot interaction}, 2006, pp. 242--249.

\bibitem{albrecht2018autonomous}
S.~V. Albrecht and P.~Stone, ``Autonomous agents modelling other agents: A
  comprehensive survey and open problems,'' \emph{Artificial Intelligence},
  vol. 258, pp. 66--95, 2018.

\bibitem{chen2020visual}
B.~Chen, S.~Song, H.~Lipson, and C.~Vondrick, ``Visual hide and seek,'' in
  \emph{Artificial Life Conference Proceedings}.\hskip 1em plus 0.5em minus
  0.4em\relax MIT Press, 2020, pp. 645--655.

\bibitem{baker2020emergent}
B.~Baker, I.~Kanitscheider, T.~Markov, Y.~Wu, G.~Powell, B.~McGrew, and
  I.~Mordatch, ``Emergent tool use from multi-agent autocurricula,'' 2020.

\bibitem{weihs2019artificial}
L.~Weihs, A.~Kembhavi, W.~Han, A.~Herrasti, E.~Kolve, D.~Schwenk, R.~Mottaghi,
  and A.~Farhadi, ``Artificial agents learn flexible visual representations by
  playing a hiding game,'' \emph{arXiv preprint arXiv:1912.08195}, 2019.

\bibitem{labash2020perspective}
A.~Labash, J.~Aru, T.~Matiisen, A.~Tampuu, and R.~Vicente, ``Perspective taking
  in deep reinforcement learning agents,'' \emph{Frontiers in Computational
  Neuroscience}, vol.~14, 2020.

\bibitem{leibfried2016deep}
F.~Leibfried, N.~Kushman, and K.~Hofmann, ``A deep learning approach for joint
  video frame and reward prediction in atari games,'' \emph{arXiv preprint
  arXiv:1611.07078}, 2016.

\bibitem{finn2017deep}
C.~Finn and S.~Levine, ``Deep visual foresight for planning robot motion,'' in
  \emph{2017 IEEE International Conference on Robotics and Automation
  (ICRA)}.\hskip 1em plus 0.5em minus 0.4em\relax IEEE, 2017, pp. 2786--2793.

\bibitem{racaniere2017imagination}
S.~Racani{\`e}re, T.~Weber, D.~Reichert, L.~Buesing, A.~Guez, D.~J. Rezende,
  A.~P. Badia, O.~Vinyals, N.~Heess, Y.~Li, \emph{et~al.},
  ``Imagination-augmented agents for deep reinforcement learning,'' in
  \emph{Advances in neural information processing systems}, 2017, pp.
  5690--5701.

\bibitem{chiappa2017recurrent}
S.~Chiappa, S.~Racaniere, D.~Wierstra, and S.~Mohamed, ``Recurrent environment
  simulators,'' \emph{arXiv preprint arXiv:1704.02254}, 2017.

\bibitem{hafner2019learning}
D.~Hafner, T.~Lillicrap, I.~Fischer, R.~Villegas, D.~Ha, H.~Lee, and
  J.~Davidson, ``Learning latent dynamics for planning from pixels,'' in
  \emph{International Conference on Machine Learning}.\hskip 1em plus 0.5em
  minus 0.4em\relax PMLR, 2019, pp. 2555--2565.

\bibitem{street2018perspective}
C.~N. Street, W.~F. Bischof, and A.~Kingstone, ``Perspective taking and theory
  of mind in hide and seek,'' \emph{Attention, Perception, \& Psychophysics},
  vol.~80, no.~1, pp. 21--26, 2018.

\bibitem{reinhold2019behavioral}
A.~S. Reinhold, J.~I. Sanguinetti-Scheck, K.~Hartmann, and M.~Brecht,
  ``Behavioral and neural correlates of hide-and-seek in rats,''
  \emph{Science}, vol. 365, no. 6458, pp. 1180--1183, 2019.

\bibitem{trafton2005enabling}
J.~G. Trafton, N.~L. Cassimatis, M.~D. Bugajska, D.~P. Brock, F.~E. Mintz, and
  A.~C. Schultz, ``Enabling effective human-robot interaction using
  perspective-taking in robots,'' \emph{IEEE Transactions on Systems, Man, and
  Cybernetics-Part A: Systems and Humans}, vol.~35, no.~4, pp. 460--470, 2005.

\bibitem{bratman2010new}
J.~Bratman, M.~Shvartsman, R.~L. Lewis, and S.~Singh, ``A new approach to
  exploring language emergence as boundedly optimal control in the face of
  environmental and cognitive constraints,'' in \emph{Proceedings of the 10th
  International Conference on Cognitive Modeling}.\hskip 1em plus 0.5em minus
  0.4em\relax Citeseer, 2010, pp. 7--12.

\bibitem{foerster2016learning}
J.~Foerster, I.~A. Assael, N.~De~Freitas, and S.~Whiteson, ``Learning to
  communicate with deep multi-agent reinforcement learning,'' in \emph{Advances
  in neural information processing systems}, 2016, pp. 2137--2145.

\bibitem{lazaridou2016multi}
A.~Lazaridou, A.~Peysakhovich, and M.~Baroni, ``Multi-agent cooperation and the
  emergence of (natural) language,'' \emph{arXiv preprint arXiv:1612.07182},
  2016.

\bibitem{grouchy2016evolutionary}
P.~Grouchy, G.~M. D’Eleuterio, M.~H. Christiansen, and H.~Lipson, ``On the
  evolutionary origin of symbolic communication,'' \emph{Scientific reports},
  vol.~6, no.~1, pp. 1--9, 2016.

\bibitem{sukhbaatar2016learning}
S.~Sukhbaatar, R.~Fergus, \emph{et~al.}, ``Learning multiagent communication
  with backpropagation,'' in \emph{Advances in neural information processing
  systems}, 2016, pp. 2244--2252.

\bibitem{mordatch2017emergence}
I.~Mordatch and P.~Abbeel, ``Emergence of grounded compositional language in
  multi-agent populations,'' \emph{arXiv preprint arXiv:1703.04908}, 2017.

\bibitem{cao2018emergent}
K.~Cao, A.~Lazaridou, M.~Lanctot, J.~Z. Leibo, K.~Tuyls, and S.~Clark,
  ``Emergent communication through negotiation,'' \emph{arXiv preprint
  arXiv:1804.03980}, 2018.

\bibitem{jain2019two}
U.~Jain, L.~Weihs, E.~Kolve, M.~Rastegari, S.~Lazebnik, A.~Farhadi, A.~G.
  Schwing, and A.~Kembhavi, ``Two body problem: Collaborative visual task
  completion,'' in \emph{Proceedings of the IEEE Conference on Computer Vision
  and Pattern Recognition}, 2019, pp. 6689--6699.

\bibitem{byravan2018se3}
A.~Byravan, F.~Lceb, F.~Meier, and D.~Fox, ``Se3-pose-nets: Structured deep
  dynamics models for visuomotor control,'' in \emph{2018 IEEE International
  Conference on Robotics and Automation (ICRA)}.\hskip 1em plus 0.5em minus
  0.4em\relax IEEE, 2018, pp. 1--8.

\bibitem{hoque2020visuospatial}
R.~Hoque, D.~Seita, A.~Balakrishna, A.~Ganapathi, A.~K. Tanwani, N.~Jamali,
  K.~Yamane, S.~Iba, and K.~Goldberg, ``Visuospatial foresight for multi-step,
  multi-task fabric manipulation,'' \emph{arXiv preprint arXiv:2003.09044},
  2020.

\bibitem{hafner2019dream}
D.~Hafner, T.~Lillicrap, J.~Ba, and M.~Norouzi, ``Dream to control: Learning
  behaviors by latent imagination,'' \emph{arXiv preprint arXiv:1912.01603},
  2019.

\bibitem{ha2018world}
D.~Ha and J.~Schmidhuber, ``World models,'' \emph{arXiv preprint
  arXiv:1803.10122}, 2018.

\bibitem{hirose2019deep}
N.~Hirose, F.~Xia, R.~Mart{\'\i}n-Mart{\'\i}n, A.~Sadeghian, and S.~Savarese,
  ``Deep visual mpc-policy learning for navigation,'' \emph{IEEE Robotics and
  Automation Letters}, vol.~4, no.~4, pp. 3184--3191, 2019.

\bibitem{wu2020spatial}
J.~Wu, X.~Sun, A.~Zeng, S.~Song, J.~Lee, S.~Rusinkiewicz, and T.~Funkhouser,
  ``Spatial action maps for mobile manipulation,'' \emph{arXiv preprint
  arXiv:2004.09141}, 2020.

\bibitem{olson2011apriltag}
E.~Olson, ``Apriltag: A robust and flexible visual fiducial system,'' in
  \emph{2011 IEEE International Conference on Robotics and Automation}.\hskip
  1em plus 0.5em minus 0.4em\relax IEEE, 2011, pp. 3400--3407.

\bibitem{hart1968formal}
P.~E. Hart, N.~J. Nilsson, and B.~Raphael, ``A formal basis for the heuristic
  determination of minimum cost paths,'' \emph{IEEE transactions on Systems
  Science and Cybernetics}, vol.~4, no.~2, pp. 100--107, 1968.

\bibitem{gao2017intention}
W.~Gao, D.~Hsu, W.~S. Lee, S.~Shen, and K.~Subramanian, ``Intention-net:
  Integrating planning and deep learning for goal-directed autonomous
  navigation,'' \emph{arXiv preprint arXiv:1710.05627}, 2017.

\bibitem{chen2018deep}
X.~Chen, A.~Ghadirzadeh, J.~Folkesson, M.~Bj{\"o}rkman, and P.~Jensfelt, ``Deep
  reinforcement learning to acquire navigation skills for wheel-legged robots
  in complex environments,'' in \emph{2018 IEEE/RSJ International Conference on
  Intelligent Robots and Systems (IROS)}.\hskip 1em plus 0.5em minus
  0.4em\relax IEEE, 2018, pp. 3110--3116.

\bibitem{shah2018follownet}
P.~Shah, M.~Fiser, A.~Faust, J.~C. Kew, and D.~Hakkani-Tur, ``Follownet: Robot
  navigation by following natural language directions with deep reinforcement
  learning,'' \emph{arXiv preprint arXiv:1805.06150}, 2018.

\bibitem{bruls2019right}
T.~Bruls, H.~Porav, L.~Kunze, and P.~Newman, ``The right (angled) perspective:
  Improving the understanding of road scenes using boosted inverse perspective
  mapping,'' in \emph{2019 IEEE Intelligent Vehicles Symposium (IV)}.\hskip 1em
  plus 0.5em minus 0.4em\relax IEEE, 2019, pp. 302--309.

\bibitem{lynch2020learning}
C.~Lynch, M.~Khansari, T.~Xiao, V.~Kumar, J.~Tompson, S.~Levine, and
  P.~Sermanet, ``Learning latent plans from play,'' in \emph{Conference on
  Robot Learning}, 2020, pp. 1113--1132.

\bibitem{dosovitskiy2015flownet}
A.~Dosovitskiy, P.~Fischer, E.~Ilg, P.~Hausser, C.~Hazirbas, V.~Golkov, P.~Van
  Der~Smagt, D.~Cremers, and T.~Brox, ``Flownet: Learning optical flow with
  convolutional networks,'' in \emph{Proceedings of the IEEE international
  conference on computer vision}, 2015, pp. 2758--2766.

\bibitem{kingma2014adam}
D.~P. Kingma and J.~Ba, ``Adam: A method for stochastic optimization,''
  \emph{arXiv preprint arXiv:1412.6980}, 2014.

\bibitem{ebert2018visual}
F.~Ebert, C.~Finn, S.~Dasari, A.~Xie, A.~Lee, and S.~Levine, ``Visual
  foresight: Model-based deep reinforcement learning for vision-based robotic
  control,'' \emph{arXiv preprint arXiv:1812.00568}, 2018.

\bibitem{juliani2018unity}
A.~Juliani, V.-P. Berges, E.~Vckay, Y.~Gao, H.~Henry, M.~Mattar, and D.~Lange,
  ``Unity: A general platform for intelligent agents,'' \emph{arXiv preprint
  arXiv:1809.02627}, 2018.

\bibitem{buesing2018learning}
L.~Buesing, T.~Weber, S.~Racaniere, S.~Eslami, D.~Rezende, D.~P. Reichert,
  F.~Viola, F.~Besse, K.~Gregor, D.~Hassabis, \emph{et~al.}, ``Learning and
  querying fast generative models for reinforcement learning,'' \emph{arXiv
  preprint arXiv:1802.03006}, 2018.

\bibitem{chung2018survey}
S.-J. Chung, A.~A. Paranjape, P.~Dames, S.~Shen, and V.~Kumar, ``A survey on
  aerial swarm robotics,'' \emph{IEEE Transactions on Robotics}, vol.~34,
  no.~4, pp. 837--855, 2018.

\bibitem{gs2018survey}
G.~GS, A.~Arockia, and V.~Berlin, ``A survey on swarm robotic modeling,
  analysis and hardware architecture,'' \emph{Procedia Computer Science,
  Science Direct}, vol. 133, pp. 478--485, 2018.

\end{thebibliography}

\end{document}